\documentclass[10pt,twocolumn,letterpaper]{article}

\usepackage{cvpr}
\usepackage{times}
\usepackage{epsfig}
\usepackage{graphicx}
\usepackage{amsmath}
\usepackage{amssymb}
\usepackage{booktabs} 
\usepackage{multirow}
\usepackage{algorithm}
\usepackage{algorithmic}
\usepackage{threeparttable}
\usepackage{caption}
\usepackage{subfigure}

\usepackage[breaklinks=true,bookmarks=false]{hyperref}

\cvprfinalcopy 

\ifcvprfinal\fi
\begin{document}

\title{Deep Saliency Hashing for Fine-grained Retrieval}

\author{Sheng Jin$^1$\hspace{0.05in}
	Hongxun Yao$^1$\hspace{0.05in}
	Xiaoshuai Sun$^1$\hspace{0.05in}
	Shangchen Zhou$^2$\hspace{0.05in}
	Lei Zhang$^3$\hspace{0.05in}
	Xiansheng Hua$^4$\\
	$^1$School of Computer Science and Technology, Harbin Institute of Technology
	$^2$SenseTime Research \\
	$^3$Hong Kong Polytechnic University, Hong Kong, China
	$^4$Damo Academy, Alibaba Group \\
	$^1${\tt\small 16B903055@stu.hit.edu.cn}
	$^1${\tt\small \{h.yao, xiaoshuaisun\}@hit.edu.cn} \\
	$^2${\tt\small zhoushangchen@sensetime.com}
	$^3${\tt\small cslzhang@comp.polyu.edu.hk}
	$^4${\tt\small xiansheng.hxs@alibaba-inc.com}
}

\maketitle

\begin{abstract}
In recent years, hashing methods have been proved to be effective and efficient for the large-scale Web media search. However, the existing general hashing methods have limited discriminative power for describing fine-grained objects that share similar overall appearance but have subtle difference. To solve this problem, we for the first time introduce the attention mechanism to the learning of fine-grained hashing codes. Specifically, we propose a novel deep hashing model, named deep saliency hashing (\textbf{DSaH}), which automatically mines salient regions and learns semantic-preserving hashing codes simultaneously. \textbf{DSaH} is a two-step end-to-end model consisting of an attention network and a hashing network. Our loss function contains three basic components, including the semantic loss, the saliency loss, and the quantization loss. As the core of \textbf{DSaH}, the saliency loss guides the attention network to mine discriminative regions from pairs of images. We conduct extensive experiments on both fine-grained and general retrieval datasets for performance evaluation. Experimental results on fine grained dataset, including Oxford Flowers-17, Stanford Dogs-120 and CUB Bird demonstrate that our \textbf{DSaH} performs the best for fine-grained retrieval task and beats strongest competitor (\textbf{DTQ}) by approximately $10\%$ on both Stanford Dogs-120 and CUB Bird. \textbf{DSaH} is also comparable to several state-of-the-art hashing methods on general datasets, including CIFAR-10 and NUS-WIDE.
\end{abstract}

\section{Introduction}
Searching for content relevant images in a large scale dataset is widely used in pratical application. Such retrieval tasks remain a challenge because of the large computational cost and the high accuracy requirement. To address the efficiency and effectiveness problems, a great number of hashing methods are proposed to map images to binary codes. These hashing methods can be classified into two categories: data-independent \cite{andoni2006Near} and data-dependent \cite{cite:CVPR18DCH, Duan2017learning}. Since data-dependent methods preserve the semantic structure of the data, they usually achieve better performance.

\begin{figure}
	\centering
	\includegraphics[width=\linewidth]{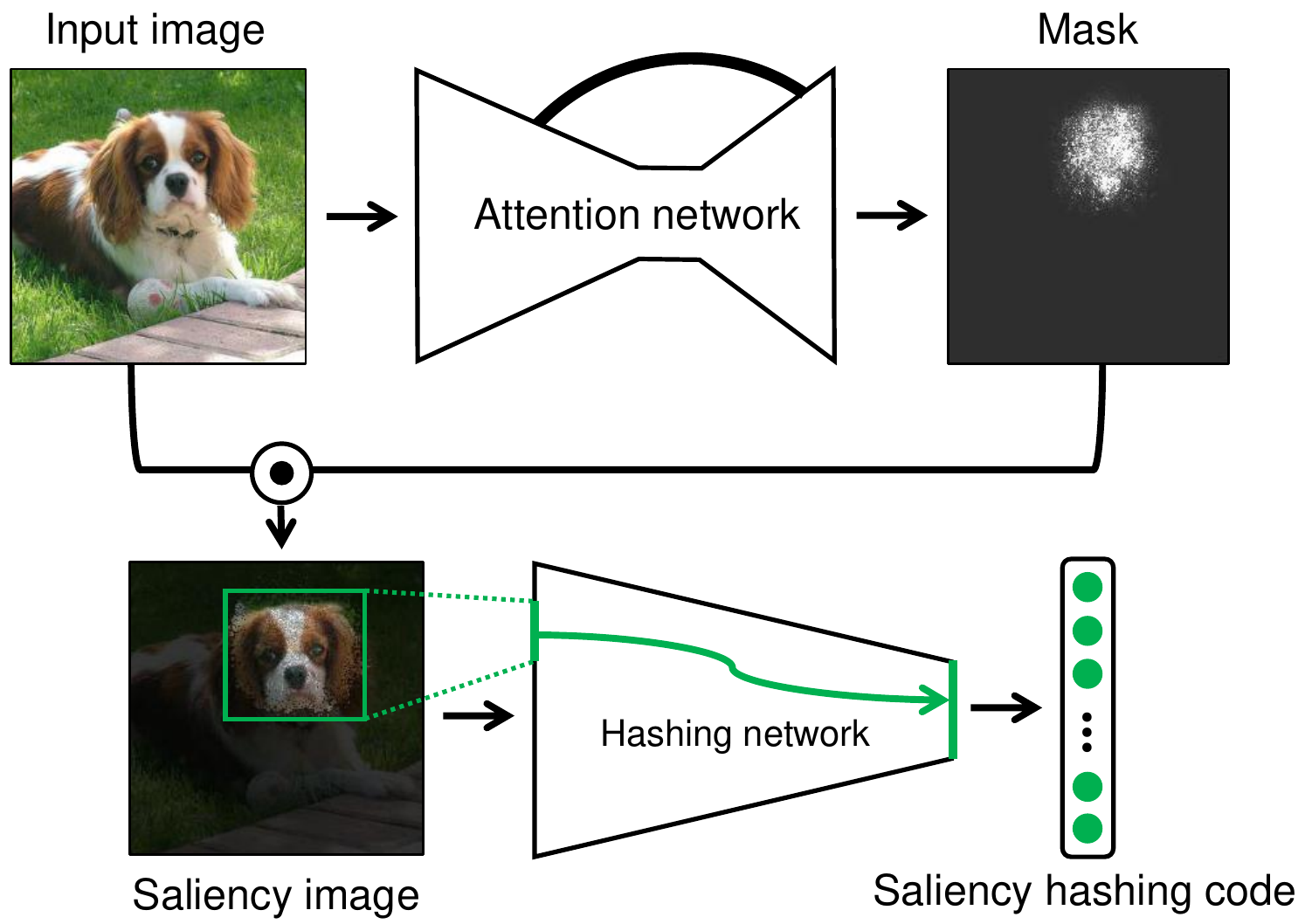}
	\caption{The main idea of our method. We propose a deep hashing method for the retrieval of fine-grained objects which share similar appearances. To produce more discriminative hashing codes, our method highlights the discriminative regions of input images using the attention network.}
	\label{Motivation:}
	\vspace{-0.5cm}
\end{figure}

Data-dependent methods can be further divided into three categories: unsupervised methods, semi-supervised methods, and supervised methods. Compared to the former two categories, supervised methods use semantic information in terms of reliable class labels to improve performance. Many representative works have been proposed along this direction, e.g., Binary Reconstruction Embedding \cite{kulis2009learning}, Column Generation Hashing \cite{li2013learning}, Kernel-based Supervised Hashing \cite{liu2012supervised}, Minimal Loss Hashing \cite{norouzi2011minimal}, Hamming Distance Metric Learning \cite{norouzi2012hamming}, and Semantic Hashing \cite{salakhutdinov2009semantic}. The success of supervised methods demonstrates that class information can dramatically improve the quality of hashing codes. However, these shallow hashing methods use hand-crafted features to represent images and generate the hashing codes. Thus, the quality of hashing codes depends heavily on feature selection, which is the most crucial limitation of such methods.

It is hard to ensure that these hand-crafted features preserve sufficient semantic information. To solve this limitation, Xia et al. \cite{xia2014supervised} introduce deep learning to hashing (\textbf{CNNH}), which performs feature learning and hashing codes learning simultaneously. Following this work, many deep hashing methods have been proposed, including Deep Cauchy Hashing \cite{cite:CVPR18DCH}, Deep Triplet Quantization\cite{liu2018deep}, Deep Supervised Hashing \cite{liu2016deep},  and Deep Semantic Ranking Hashing \cite{zhao2015deep}. Extensive experiments demonstrate that deep hashing methods achieve significant improvements.

Nevertheless, existing deep hashing methods are mostly studied and validated on general datasets, e.g., CIFAR-10 \cite{krizhevsky2009learning} and NUS-WIDE \cite{chua2009nus}. These datasets contain only a few categories with a large number of images per class. Besides, different classes have significant differences in appearance which make the problem simpler than real-world cases. To support practical applications, two crucial issues still need to be considered. Firstly, a robust hashing method should be able to \textit{distinguish fine-grained objects}. The major challenge is that these fine-grained objects share similar overall appearance, making the inter-class differences more important than the intra-class variance. Secondly, hashing methods should be able to support \textit{a large number of categories}. Different from CIFAR-10, the existing fine-grained dataset consists of much more categories with small amounts of images per class. This raises another challenge which is how to generate hashing codes for much more categories with relatively fewer data. 

Similar challenges exist in other fields of computer vision, e.g., fine-grained classification \cite{wang2016mining} and person re-id \cite{zhao2017deeply}. In these fields, representative methods deal with the above challenges by mining discriminative parts for each category, either manually \cite{wang2016mining} or automatically \cite{fu2017look,zhao2017deeply}. The deep methods proposed by \cite{fu2017look} and \cite{zhao2017deeply} automatically mine salient regions and achieve remarkable improvements compared to traditional state-of-the-arts. Inspired by such methods, we propose a novel deep hashing method for fine-grained retrieval, termed Deep Saliency Hashing (\textbf{DSaH}), to solve the two above-mentioned challenges jointly. 

The main idea of the proposed saliency hashing is depicted in Fig.~\ref{Motivation:}. Specifically, \textbf{DSaH} is an end-to-end CNN model that simultaneously mines discriminative parts and learns hashing codes. \textbf{DSaH} contains three components. The first component, named attention module, is a full convolutional network aiming to generate a saliency image from the original image. The second component, named hashing module, is a deep network based on VGG-16 model aiming to map images to hashing codes. And the third component is a loss function. The loss function contains 1) a novel semantic loss to measure the semantic quality of hashing codes based on the pairwise labels; 2) a saliency loss of image quadruples to mine discriminative region; and 3) a quantization loss to learn the optimal hashing codes from the binary-like codes. All the components are intergraded into a unified framework. We iteratively update the attention network and the hashing network to learn better hashing codes. The main contributions of \textbf{DSaH} are three-fold:
\begin{itemize}
	\item We propose a deep hashing method by integrating saliency mechanism into hashing. To the best of our knowledge, \textbf{DSaH} is the first attentional deep hashing model specially designed for fine-grained tasks.
	
\item A novel saliency loss of image quadruples is proposed to guide the attention network for automatic discriminative regions mining. Experimental results demonstrate that the fine-grained categories can be better distinguished based on these attractive regions.

\item Experimental results on both general hashing datatsets and fine-grained retrieval datasets demonstrate the superior performance of our method in comparison with many state-of-art hashing methods.

\end{itemize}
\begin{figure*}
	\centering
	\includegraphics[width=0.9\linewidth]{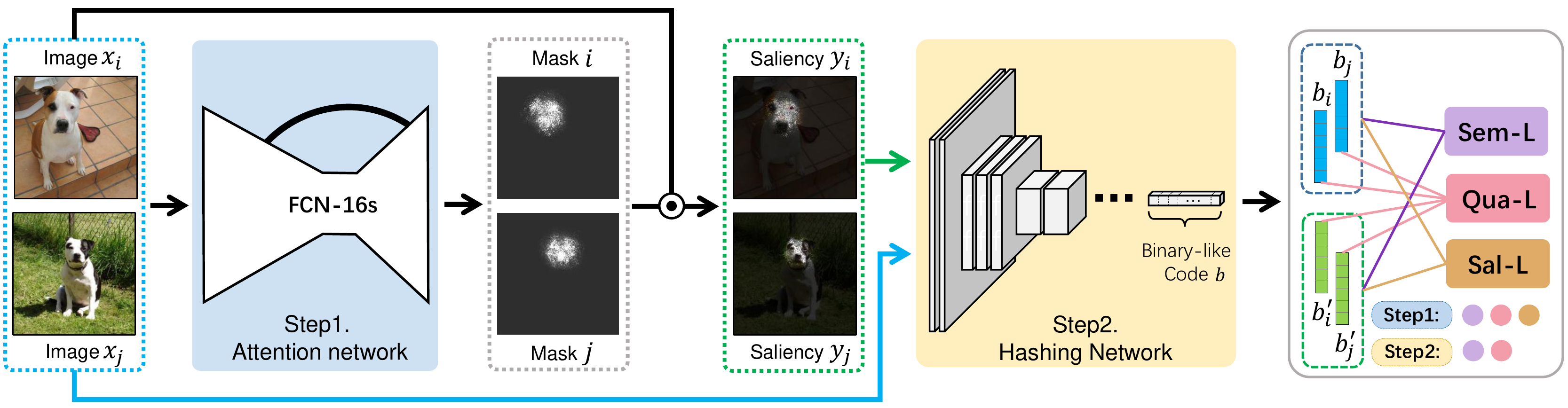}
	\caption{The proposed framework for deep saliency hashing (\textbf{DSaH}). \textbf{DSaH} is comprised of three components: (1) an attention network based on fcn-16s for learning a saliency image, (2) a hashing network based on vgg-16 for learning hashing codes, (3) a set of loss functions including a semantic loss (Sem-L), a saliency loss (Sal-L) and a quantization loss (Qua-L) for optimization. In the training stage, the attention network and the hashing network are trained iteratively to mine discriminative regions (Step1) and learn semantic-preserving hashing codes (Step2). For attention network, the whole set of loss functions are used. For hashing network, we only use the semantic loss and quantization loss.}
	\label{Fig:framework}
\end{figure*}

\section{Related Work}
We introduce the most related works from two aspects: \textit{hashing code learning} and \textit{discriminative part mining}.

{\bfseries Deep Hashing}
In the last few years, deep convolutional neural network \cite{xia2014supervised} has been employed to supervised hashing. Followed by \cite{xia2014supervised}, some representative works have been proposed \cite{li2015feature,lin2015deep,liu2016deep,zhao2015deep}. These deep methods are proved to be effective for general object retrieval, where different categories have significant visual differences (e.g., CIFAR-10). Recently, deep hashing methods emerge as a promising solution for efficient person re-id \cite{zhang2015bit,zhu2017part}. Different from object retrieval, human bodies share similar appearance with subtle difference in some salient regions. Zhang \cite{zhang2015bit} et al. propose DRSCH and introduce hashing into person re-id. DRSCH is a triplet-based model and encodes the entire person image to hashing codes without considering the part-level semantics. PDH \cite{zhu2017part} integrates the part-based model into the triplet model and achieves significant improvements. However, the part partition strategy of PDH is specified based on human structure. Since there are huge variations in scale, multiple instances etc, in typical fine-grained datasets (e.g., CUB Bird \cite{WahCUB_200_2011}), the part partitioning strategy of PDH is not suitable for non-human fine-grained objects. In this paper, we introduce the hashing method into fine-grained retrieval, where an attention network is embedded to mine salient regions for accurate code learning.

\indent {\bfseries Mining Discriminative Regions} The key challenge of learning accurate hashing codes for fine-grained objects is to locate the discriminative regions in images. Facing a similar challenge, to boost the performance of fine-grained image classification, researchers have proposed various salient region localization approaches \cite{singh2012unsupervised,xu2013human,zheng2017pose}. Previous works locate the salient regions either by unsupervised methods \cite{liu2016fully,xiao2015application} or by leveraging manual part annotations \cite{zhang2016spda,zhang2014part}. Following these works, the recent hashing methods \cite{bai2017deep,shen2017deep} locate salient regions to improve performance in an unsupervised manner. DPH \cite{bai2017deep} uses GBVS \cite{harel2007graph} to calculate the saliency scores for each pixel. Then a series of salient regions are generated by increasing the threshold values. Shen et al. propose a cross-model hashing method, named TVDB \cite{shen2017deep}, which adopts RPN \cite{ren2015faster} to detect salient regions and encodes the regional information of image, the semantic dependencies, as well as the cues between words by two modal-specific networks. However, these hashing methods use the off-the-shelf models to locate salient regions, which might not be accurate for new images or specific tasks. Instead, our model trains a saliency prediction network jointly with the hashing network, where the two modules are optimized together toward a unified objective.  

Recent methods \cite{liu2016fully,zhao2016diversified} try to discover discriminative regions automatically by deep networks. These deep methods do not require labeling information, such as the labeled part masks or boxes, but only use the class label information. Zhao et al. \cite{zhao2017deeply} use the similarity information (a pair of person images about the same person or not) to train part model specially for person matching. Similar intuitions can be found in recent fine-grained classification methods. Fu et al. \cite{fu2017look} propose a novel recurrent attention convolutional neural network, named RA-CNN, to discover salient regions and learn region-based feature representation recursively. The basic idea of this method is that salient region localization and fine-grained feature learning are mutually correlated and thus can reinforce each other. Motivated by \cite{fu2017look,zhao2017deeply}, we adopt a novel attention network to automatically mine the salient region for learning better hashing codes. To the best of our knowledge, it is the first time that attention mechanism is formally employed to fine-grained hashing.

\section{Deep Saliency Hashing}
Fig.~\ref{Fig:framework} shows the proposed \textbf{DSaH} framework. Our method includes three main components. The first component is an attention network. The attention network maps each input image into a saliency image. The second component is a hashing network. The hashing network learns binary-like codes from an original image and its saliency image. The third component is a set of loss terms, including pairwise semantic loss, saliency loss of image quadruples, and quantization loss. The semantic loss requires the hashing codes learned from each image pair to preserve semantic information. The saliency loss guides the attention network to highlight the salient regions of original images. The quantization loss is devised to measure the loss between binary-like codes and hashing codes after binarization. The whole cost function is written as below:

\begin{equation}
J=J_{sem}+J_{sal}+J_{reg},
\end{equation}
where $J_{sem}$ represents the semantic loss, $J_{sal}$ represents the saliency loss and $J_{reg}$ represents the quantization loss.

\subsection{The Attention Network}
The attention network is proposed to map the original image $x_i$ to the saliency image $y_i$ : $x_i \rightarrow y_i$. This module includes two stages. In the first stage, we assign a saliency value of each pixel in the original image. Then we obtain the saliency image by highlighting the salient pixels.

As described above, a dense prediction problem needs to be solved in the first stage. The location $(p,q)$ in image $x_i$ is denoted as ${x_i}(p,q)$. We denote the learned saliency value of each pixel in image $x_i$ as ${Salmap_i}(p,q)$. Long. et al \cite{Long_2015_CVPR} prove that FCN is effective for dense prediction which maps each pixel of an image to a label vector.

Motivated by FCN \cite{Long_2015_CVPR}, we propose an FCN-based attention network, as illustrated in Fig.~\ref{Fig:framework}, to discover the salient region automatically. Different from semantic segmentation approaches, our method does not predict a label vector but assign a saliency value for each pixel.

In the first stage, the proposed FCN-based attention network maps each pixel of images to a saliency value:
\begin{equation}
Salmap_i = Attention(x_{i}).
\end{equation}
%
To regularize the output, we further normalize the saliency map so its value is between 0 and 1:

\begin{equation}
{Salmap_i}(p,q) = \frac{{Salmap_i}(p,q)-\min (Salmap_i)}{\max (Salmap_i) -\min (Salmap_i)}.
\end{equation}

Then the saliency image $y_i$ is computed through a matrix dot product by the original image and their saliency map:

\begin{equation}
y_i = {Salmap_i} \odot x_i.
\end{equation}

Then we encode the saliency image $y_i$ by the hashing network. We can obtain the saliency loss defined in Eq.~\ref{eq:attention}. By iteratively updating the parameters with the saliency loss, the attention network is gradually fine-tuned to mine discriminative regions automatically.
\subsection{The Hashing Network}
As shown in Fig.~\ref{Fig:framework}, We directly adopt a pre-trained VGG-16 \cite{simonyan2014very} as the base model of the hashing network. The raw image pixel, from either the original image and the saliency image, is the input of the hashing model. The output layer of VGG is replaced by a hashing layer where the dimension is defined based on the length of the required hashing code. The hashing network is trained by the semantic loss (Eq.\ref{eq:semanticlossori}) and quantization loss (Eq.~\ref{eq:quantizationloss}).
\subsection{Loss Functions}
\textbf{Semantic Loss} Similar to other hashing methods, our goal is to learn efficient binary codes for images: $x\rightarrow b\in{\{1,-1\}}^k$, where $k$ denotes the number of hashing bits. Since discrete optimization is difficult to be solved by deep networks, we firstly ignore the binary constraint and concentrate on binary-like code for network training, where the binary-like code is denoted as $\mu_i$. Then we obtain the optimal hashing codes $b_i$ from $\mu_i$. We define a pairwise semantic loss to ensure the binary-like codes preserve relevant semantic information. Since each image in the dataset owns a unique class label, image pairs could be further labeled as similar or dissimilar:
\begin{equation}
\label{eq:similarity}
S_{ij}=
\begin{cases}
1 & \text{images $x_i$ and $x_j$ share same class label}\\
0 & \text{otherwise},\\
\end{cases}
\end{equation}
where $S_{ij}$ denotes the pairwise label of images $x_i$, $x_j$. To preserve semantic information, the binary-like codes of similar images should be as close as possible while the binary-like codes of dissimilar images should be as far as possible. Since hashing methods select the hamming distance to measure similarity in the testing phase, we use $\mu_{i}^T*\mu_{j}$ to calculate the distance of image $x_i$ and $x_j$. Given $b_i \in {\{1,-1\}}$, the inner product of $\mu_{i}$ and $\mu_{j}$ is in the range of $(-k,k)$. Thus we adopt $\frac{\mu_{i}^T* \mu_{j}+k}{2k}$ to transform the inner product to $(0,1)$. The result of such linear transformation is regarded as an estimation of the pairwise label $S_{i,j}$. The semantic loss of original image pairs is written as below:

\begin{equation}
\label{eq:semanticlossori}
J_{sem\_ori}=\sum_{i,j} {{\left \Vert S_{i,j}-\frac{ \mu_{i}^T* \mu_{j}+k}{2k}\right \Vert}}_{2}.
\end{equation}

Our method also requires hashing codes to learn semantic information from the salient region. To achieve this objective, the hashing codes of the saliency image also need to preserve semantic information. Specifically, we use the attention network to map the original image pairs ($x_i$, $x_j$) into saliency image pairs ($y_i$, $y_j$). Similar to Eq.~\ref{eq:semanticlossori}, the semantic loss of saliency image pairs is defined as below:
\begin{equation}
\label{eq:semanticlosssal}
J_{sem\_sal}=\sum_{i,j} {{\left \Vert S_{i,j}-\frac{ {\mu'}_{i}^T* {\mu'}_{j}+k}{2k}\right \Vert}}_{2}.
\end{equation}

To learn the saliency image, we propose an attention network. The key idea is that the hashing codes learned from saliency images are more discriminative. A saliency loss is defined to guide the attention model to highlight the salient regions of the original image. Firstly, the proposed attention network outputs the saliency image $y_i$ from the original image $x_i$ : $x_i \rightarrow y_i$. Then the original image $x_i$ and its saliency image $y_i$ are mapped to binary-like codes by the hashing model, which is denoted as $\mu_i$, $\mu'_i$ respectively.

\textbf{Saliency Loss} Similar to the semantic loss, we use image pairs to define the saliency loss. The original images $x_i$ and $x_j$ are taken as the original image pair. Their saliency images $y_i$ and $y_j$ are regarded as the saliency image pair. The original image pair and their saliency image pair construct an image quadruple. The binary-like codes of saliency image $\mu'_i$ and $\mu'_j$ are more similar or dissimilar than those of the original image pair $\mu_i$ and $\mu_j$ according to whether images $x_i$ and $x_j$ share the same labels or not. Eq.~\ref{eq:semanticlossori}, $\frac{\mu_{i}^T* \mu_{j}+k}{2k}$ is used to approximate the value of pairwise label. We denote distance of the pairwise label and the estimated value from original (saliency) image pairs as $d_{ij}$ ($d'_{ij}$):

\begin{equation}
d_{ij}= {{\left \Vert S_{i,j}-\frac{ \mu_{i}^T* \mu_{j}+k}{2k}\right \Vert}}_{2}, d'_{ij}= {{\left \Vert S_{i,j}-\frac{ {\mu'}_{i}^T* \mu'_{j}+k}{2k}\right \Vert}}_{2}.
\end{equation}
As described before, the saliency loss with respect to the quadruples of images is written as:

\begin{equation}
\label{eq:saliencyloss}
J_{sal}=\sum_{i,j} \max (m-d_{ij}+d'_{ij},0),
\end{equation}
where $m>0$ is a margin threshold. When the value of $d_{ij}-d'_{ij}$ is below the margin threshold $m$, the saliency loss punishes the attention network to make it better highlight the salient regions.

\textbf{Quantization Loss} The binary constraint of hashing codes makes it intractable to train an end-to-end deep model with backpropagation algorithm. As discussed in \cite{liu2016deep}, some widely-used relaxation scheme working with non-linear functions, such as sigmoid function,  would inevitably slow down or even restrain the convergence of the network \cite{krizhevsky2012imagenet}. To overcome such limitation, we adopt a similar regularizer on the real-valued network outputs to approach the desired binary codes. The quantization loss is written as:

\begin{equation}
\label{eq:quantizationloss}
J_{reg}=\sum_{i} {{\left \Vert \mu_i-b_i\right \Vert}}_{1}+{{\left \Vert \mu'_i-b'_i\right \Vert}}_{1},
\end{equation}
where $b\in{\{1,-1\}}^k$. Since $b_i$ only appears in the quantization loss, we minimize this loss to obtain the optimal hashing codes. Obviously, the sign of $b_i$ should be same as that of the binary-like codes $\mu_i$. Thus the hashing codes $b_i$ can be directly optimized as:
\begin{equation}
\label{eq:quantized}
b_i= sign(\mu_i).
\end{equation}

\subsection{Alternating Optimization}
The overall training objective of \textbf{DSaH} integrates the pairwise semantic loss defined in Eq.~\ref{eq:semanticlossori}, the saliency loss of image quadruples defined in Eq.~\ref{eq:saliencyloss} and the quantization loss defined in Eq.~\ref{eq:quantizationloss}. \textbf{DSaH} is a two-stage end-to-end deep model which consists of an attention network, i.e. $Attention$, for automatic saliency image estimation and a shared hashing network, i.e. $Hash$, for discriminative hashing codes generation. As shown in Algorithm.~\ref{alg:DSaH}, we train the attention network and the hashing network iteratively.

In particular, for the shared hashing model, we update its parameters according to the following overall loss:

\begin{equation}
\label{eq:hashing}
\begin{aligned}
J_{hashing}=&\sum_{i,j} {\lambda J_{sem\_ori}(x_i,x_j) + \lambda J_{sem\_sal}(y_i,y_j)} \\
&+\sum_{i}{J_{reg\_ori}(x_i)+J_{reg\_sal}(y_i)}.
\end{aligned}
\end{equation}
By minimizing this term, the shared hashing model is trained to preserve the relative similarity in of original image pairs and that of saliency image pairs.

The attention network is trained by the following loss:
\begin{equation}
\label{eq:attention}
\begin{aligned}
J_{attention}=&\sum_{i,j}{\alpha J_{sal}(x_i,x_j,y_i,y_j) + \lambda J_{sem\_sal}(y_i,y_j)}\\
&+\sum_{i}{J_{reg\_sal}(y_i)}.
\end{aligned}
\end{equation}
By minimizing this term, the attention network is trained to mine salient and semantic-preserving regions of the input image, leading to more discriminative hashing codes.
\renewcommand{\algorithmicrequire}{ \textbf{Input:}}
\renewcommand{\algorithmicensure}{ \textbf{Output:}}
\begin{algorithm}[tb]
	\caption{Deep Saliency Hashing}
	\label{alg:DSaH}
	\begin{algorithmic}[1]
		\REQUIRE  Training set and their corresponding class label, Total epochs $T$ of deep optimization;
		\ENSURE  Hashing function: $Hash(x|\Theta_{2})$ Attention function: $Attention(x|\Theta_{1})$.
		\STATE For the entire training set, construct the pairwise label matrix $S$ according to Eq.~\ref{eq:similarity}.
		\FOR{$t=1:T epoch$}
		\STATE Compute $B$ accordi.ng to Eq.~\ref{eq:salienytest}
		\STATE Update $\Theta_{1}$ according to Eq.~\ref{eq:attention}
		\STATE Update $\Theta_{2}$ according to Eq.~\ref{eq:hashing}
		\ENDFOR
		\RETURN $Hash(x|\Theta_{2})$, $Attention(x|\Theta_{1})$.
	\end{algorithmic}
\end{algorithm}

\subsection{Out-of-Sample Extension}
After the model is trained, we can use it to encode an input image with a $k$-bit binary-like code. Since the deep saliency hashing model consists of two networks, firstly the image $x_i$ is mapped to the salient image $y_i$:
\begin{equation}
\label{eq:salienytest1}
y_i=Attention(x_i).
\end{equation}

Then the hashing networks map $y_i$ to binary-like codes:
\begin{equation}
\label{eq:salienytest2}
\mu_i=Hash(y_i).
\end{equation}

As discussed in Quantization Loss according to Eq.~\ref{eq:quantized}, we adopt the sign function to produce the hashing codes
\begin{equation}
\label{eq:salienytest}
b_{i}=sign(\mu_{i})=sign(Hash(Attention(x_{i}))).
\end{equation}

%


\section{Experiments}
In order to test the performance of our proposed \textbf{DSaH} method, we conduct experiments on two widely used image retrieval datasets, i.e. CIFAR-10 and NUS-WIDE, to verify the effectiveness of our method for the general hashing task. Then we conduct experiments on three fine-grained datasets: Oxford Flower-17, Stanford Dogs-120 and CUB Bird to prove that (1) the discriminative region of images can improve retrieval performance of hashing codes on fine-grained cases, and (2) the attention model can effectively mine the saliency region of images. 

\subsection{Dataset and Evaluation Metric}
{\bfseries CIFAR-10}\cite{krizhevsky2009learning} consists of 60000 32$\times$32 images in 10 classes. Each image in dataset belongs to one class (6000 images per class). We randomly select 100 images per class as the test set and 500 images per class from the remaining images as the training set.

{\bfseries NUSWIDE}\cite{chua2009nus} is a multi-label dataset, including nearly 270k images with 81 semantic concepts. Followed \cite{liu2011hashing} and \cite{xia2014supervised}, we select the 21 most frequent concept. Each of concepts is associated with at least 5000 images. We sample 100 images from each concept to form a test set and 500 images per class from the rest images to form a train set. 

{\bfseries Oxford Flower-17}\cite{khosla2011novel} dataset consists of 1360 images belonging to 17 mutually classes. Each class contains 80 images. The dataset is divided into three parts, including a train set, test set, and validation set, with 40 images, 20 images, and 20 images respectively. We combine the validation set and train set to form the new training set.

{\bfseries Stanford Dogs-120}\cite{nilsback2006visual} consists of 20,580 images in 120 classes. Each class contains about 150 images. The dataset is divided into: the train set (100 images per class) and test set (totally 8580 images for all categories).

{\bfseries CUB Bird}\cite{WahCUB_200_2011} includes 11,788 images in mutually 200 classes. The dataset is divided into: the train set(5794 images) and the test set(5994 images).

We mainly use Mean Average Precision (\textbf{MAP})and Precision-Recall curves for quantitative evaluation.
\subsection{Comparative Methods}
For the general datasets, including CIFAR-10 and NUSWIDE dataset, we compare our method (\textbf{DSaH}) with six deep hashing method: CNNH \cite{xia2014supervised}, DNNH \cite{lai2015simultaneous}, DSH \cite{liu2016deep}, DQN \cite{cao2016deep}, DVSQ \cite{cao2017deep}, DPSH \cite{li2015feature} and three shallow methods: ITQ-CCA \cite{gong2013iterative}, KSH \cite{liu2012supervised}, SDH \cite{shen2015supervised}. For shallow hashing method, we use deep features extracted by VGG-16 to represent an image. For fair comparison, we replace the VGG-F network used as base model in DPSH \cite{li2015feature} which achieves the best retrieval performance in comparative methods, with the VGG-16 model, named DPSH++. We mainly compared with DPSH++.

For the fine-grained datasets, our method (\textbf{DSaH}) is compared with five deep methods: DSH \cite{liu2016deep}, DQN \cite{cao2016deep}, DPSH \cite{li2015feature}, DCH \cite{cite:CVPR18DCH}, DTQ \cite{liu2018deep} and four shallow method: SDH \cite{shen2015supervised}, LFH \cite{zhang2014supervised}, KSH \cite{liu2012supervised}, FastH \cite{lin2014fast}. For fair comparison, firstly we finetune VGG-16 on each fine grained dataset respectively for classification, respectively. Then these non-deep hashing methods use CNN features extracted by the output of the second full-connected layer (fc7) in the finetuned VGG-16 network. To be more fair, we replace the base model(vgg16) of our method with alexnet, which is same as DCH \cite{cite:CVPR18DCH} and DTQ \cite{liu2018deep}, named \textbf{DSaH-}.

\begin{table*}[tp]
	\small
	\vspace{-0.0cm}
	\centering
	\caption{Mean Average Precision (MAP) results for different number of bits on general datasets}
	\label{tab1:}
	\begin{tabular}{r|cccc|cccc}
		\hline
		\multirow{2}{*}{Dataset}&
		\multicolumn{4}{c|}{CIFAR-10}&\multicolumn{4}{c}{NUSWIDE}\cr\cline{2-9}
		&12 bits&24 bits&36 bits&48 bits&12 bits&24 bits&36 bits&48 bits\cr
		\hline
		ITQ-CCA\cite{gong2013iterative}&0.435&0.435&0.435&0.435&0.526&0.575&0.572&0.594\cr
		KSH\cite{liu2012supervised}&0.556&0.572&0.581&0.588&0.618&0.651&0.672&0.682\cr
		SDH\cite{shen2015supervised}&0.558&0.596&0.607&0.614&0.645&0.688&0.704&0.711\cr
		CNNH\cite{xia2014supervised}&0.439&0.476&0.472&0.489&0.611&0.618&0.628&0.608\cr
		DNNH\cite{lai2015simultaneous}&0.552&0.566&0.558&0.581&0.674&0.697&0.713&0.715\cr
		DPSH\cite{li2015feature}&0.713&0.727&0.744&0.757& 0.794& 0.822&0.838&0.851\cr	
		DQN\cite{cao2016deep}&0.554&0.558&0.564&0.580&0.768&0.776&0.783&0.792\cr
		DSH\cite{liu2016deep}&0.6157&0.6512&0.6607&0.673&0.695&0.713&0.732&0.6755\cr
		DVSQ\cite{cao2017deep}&0.715&0.730&0.749&0.760&0.788&0.792&0.795&0.803\cr
		\hline
		DPSH++\cite{li2015feature}&0.7834&0.8183&0.8294&0.8317&0.8271&0.8508&0.8592&0.8649\cr			
		\textbf{DSaH}&{\bf 0.8003}&{\bf 0.8457}&{\bf 0.8476}&{\bf 0.8478}&{\bf 0.838}&{\bf 0.854}&{\bf 0.864}&{\bf 0.873} \cr
		\hline
	\end{tabular}
\end{table*}
\begin{table*}[htb]
	\small
	\vspace{-0.3cm}
	\centering
	\caption{\textbf{MAP} results for different number of bits on three fine-grained datasets}
	\label{tab2:}
	\begin{tabular}{r|cccc|cccc|cccc}
		\hline
		\multirow{2}{*}{Dataset}&
		\multicolumn{4}{c|}{Oxford Flower-17}&\multicolumn{4}{c|}{Stanford Dogs-120}&\multicolumn{4}{c}{CUB Bird}\cr\cline{2-13}
		&12 bits&24 bits&36 bits&48 bits&12 bits&24 bits&36 bits&48 bits&12 bits&24 bits&36 bits&48 bits\cr
		\hline
		DQN\cite{cao2016deep}&0.476&0.537&0.562&0.573&0.0089 & 0.0127&0.0347 &0.0531&-&-&-&-\cr
		DSH\cite{liu2016deep}&0.566&0.614&0.637&0.680&0.0119 &0.0115&0.0117&0.0119&0.0108&0.0107&0.0108&0.0109\cr
		DCH\cite{cite:CVPR18DCH}&0.9023&0.9117&0.9449&0.9534&0.0287&0.1971&0.3090&0.3073&0.0198&0.0725&0.1112&0.1676\cr
		DTQ\cite{liu2018deep}&0.9077&0.9155&0.9203&0.9324&0.0253&0.0273 &0.0268&0.0271&0.0198&0.0233&0.0241&0.0228\cr
		\hline
		\textbf{DSaH-}&{\bf 0.9273}&{\bf 0.9354}&{\bf 0.9471}&{\bf 0.9565}&{\bf 0.2442}&{\bf 0.2874}&{\bf 0.3628} &{\bf 0.4075} &{\bf 0.0912}&{\bf 0.2087}&{\bf 0.2318}&{\bf 0.2847}\cr
		\hline
		SDH\cite{shen2015supervised}&0.1081&0.1399&0.1169&0.1446&0.0091&0.0176&0.090&0.0365&0.0148&0.0151&0.0154&0.0156\cr
		LFH\cite{zhu2016deep}&0.1887&0.4755&0.6363&0.8137&0.0249&0.0247&0.0211&0.0244&0.0064&0.0064&0.0065&0.0067\cr
		KSH\cite{liu2012supervised}&0.2431&0.5012&0.2530&0.3553&0.0136&0.1228&0.1343&0.1930&-&-&-&-\cr
		FastH\cite{lai2015simultaneous}&0.4018&0.5244&0.5281&0.5355&0.0434&0.2231&0.3643&0.3927&0.0228&0.0372&0.0423&0.0564 \cr
		DPSH++\cite{li2015feature}&0.6578&0.8295&0.8605&0.8982&0.2778 &0.4409 & 0.5054&0.5247 &0.0723&0.0764&0.0838&0.0792\cr
		\hline		
		\textbf{DSaH}&{\bf 0.9325}&{\bf 0.9467}&{\bf 0.9692}&{\bf 0.9756}&{\bf 0.3976}&{\bf 0.5283}&{\bf 0.5950} &{\bf 0.6452} &{\bf 0.1408}&{\bf 0.2817}&{\bf 0.3428}&{\bf 0.4313}\cr	
		\hline
	\end{tabular}
\end{table*}

\subsection{Implementation Details}
The \textbf{DSaH} method is implemented based on PyTorch and the deep model is trained by batch gradient descent. As shown in Fig.~\ref{Fig:framework}, our model consists of an attention network and a hashing model. We use VGG-16 as the base model. It worth mentioning that VGG-16 is not finetuned on each dataset. The full convolutional network \cite{Long_2015_CVPR} is adopted as the base model for the attention network. As discussed in \cite{Long_2015_CVPR}, FCN is improved with multi-resolution layer combinations. We use the fusing method of FCN-16s to improve performance. Practically, we train the attention network before the hashing network. If we first train the hashing network, the attention network might output a semantic-irrelevant saliency image, which would be a bad sample and guide the training of hashing model to a wrong direction.
%

Since we propose a novel attention model for hashing, we conduct additional experiments on three fine-grained datasets, \textbf{Oxford Flower-17}, \textbf{Stanford Dogs-120} and \textbf{CUB Bird} to further prove its effectiveness. Finally, we also show some typical examples of the saliency images learned by proposed attention network.

Additionally, we conduct analytical experiments to discuss these problems: (1)the analysis of hyper-parameters, (2)the convergence of the two networks, (3)the effectiveness of each loss. (4)the learned salient region

{\bfseries Network Parameters} In our method, the value of hyper-parameter $\lambda$ is 30 and $\alpha$ is 40. We use the mini-batch stochastic gradient descent with 0.9 momentum. We set the value of the margin parameters $m$ as $k/4$, where $k$ is the bits of hashing codes. The mini-batch size of images is fixed as 32 and the weight decay parameter as 0.0005.

\subsection{Experimental Results for Retrieval}
{\bfseries Performance on general hashing datasets} The Mean Average Precision (MAP,\%) results of different methods for different numbers of bits on NUSWIDE and CIFAR-10 dataset are shown in TABLE~\ref{tab1:}. Experimental results on CIFAR-10 dataset show that \textbf{DSaH} outperforms existing best retrieval performance (DPSH \cite{liu2016deep}) by $8.73\%$, $9.13\%$, $11.87\%$, $9.08\%$ correspond to different hash bits. Similar to the other hashing methods, we also conduct experiments for large-scale image retrieval. For NUSWIDE dataset, we follow the setting in \cite{liu2011hashing} and \cite{xia2014supervised}, and if two images share at least one same label, they are considered same. The experimental results of NUSWIDE dataset on TABLE~\ref{tab1:} show that our proposed method outperforms the best retrieval baseline (DPSH \cite{liu2016deep}) by $4.4\%$, $3.2\%$, $2.6\%$, $2.2\%$. According to the experimental results, \textbf{DSaH} can be clearly seen to be more effective for traditional hashing task.To ensure fairness, we conduct experiments on different hashing methods based on the same base model. The experimental results shown in TABLE~\ref{tab1:} prove that our method still outperform DPSH++ by $1.69\%$, $2.74\%$, $1.82\%$, $1.61\%$ on NUSWIDE dataset and by $1.09\%$, $0.31\%$, $0.46\%$, $0.72\%$ on CIFAR-10 dataset. 

%

\begin{figure}
	\begin{center}
		\includegraphics[width=0.9\linewidth]{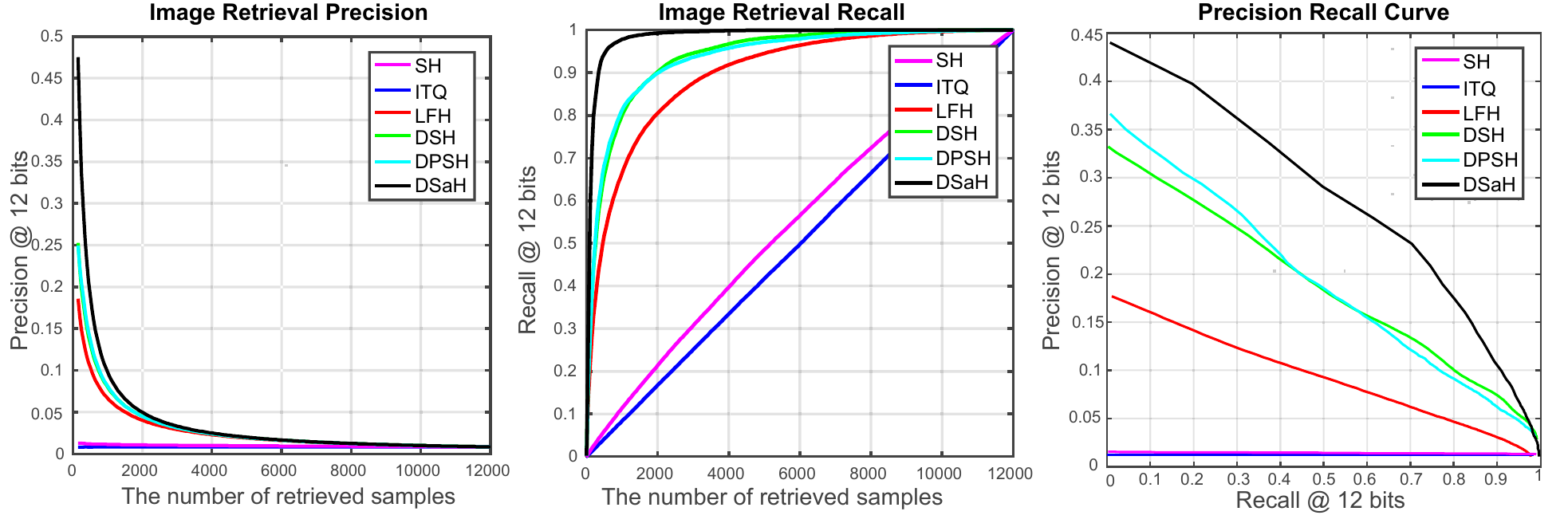}
	\end{center}
	\caption{Comparison of retrieval performance of DSH method and the other hashing methods on Stanford Dogs}
	\label{fig:stanford}
	\vspace{-0.3cm}
\end{figure}

{\bfseries Performance on fine-grained datasets} The MAP results of different methods on fine-grained datasets are shown in TABLE~\ref{tab2:}. The precision curves are shown in  Fig.~\ref{fig:stanford}. Results on Oxford Flower-17 show that \textbf{DSaH} outperforms existing best retrieval performance by a very large margin $2.48\%$, $3.12\%$, $4.43\%$, $2.22\%$ correspond to different hash bits. We also conduct experiments on a large fine-grained dataset. For Stanford dog-120 and CUB Bird, this dataset contains more categories and has smaller inter-class variations across different classes. The MAP results of all methods on these datasets are listed in TABLE ~\ref{tab2:} which show that the proposed \textbf{DSaH} method substantially outperforms all the comparison methods. \textbf{DSaH} achieves absolute increases of $11.98\%$, $12.74\%$, $8.97\%$, $12.04\%$ and $6.85\%$, $20.53\%$, $23.16\%$, $26.37\%$. To ensure fairness, \textbf{DSaH-} use the same base-model as DTQ\cite{liu2018deep} and DCH\cite{cite:CVPR18DCH}. {\bf DSaH-} still outperfoms these methods by about 10\% on Stanford dog-120 and CUB Bird datasets. Compared with the MAP results on traditional hashing task, our method is proved to achieve a significant improvement in fine-grained retrieval.


\subsection{Exploration Experiment}
\begin{figure}   
    	\begin{minipage}[t]{0.66\linewidth} 
    		\centering   
    		\includegraphics[width=2in]{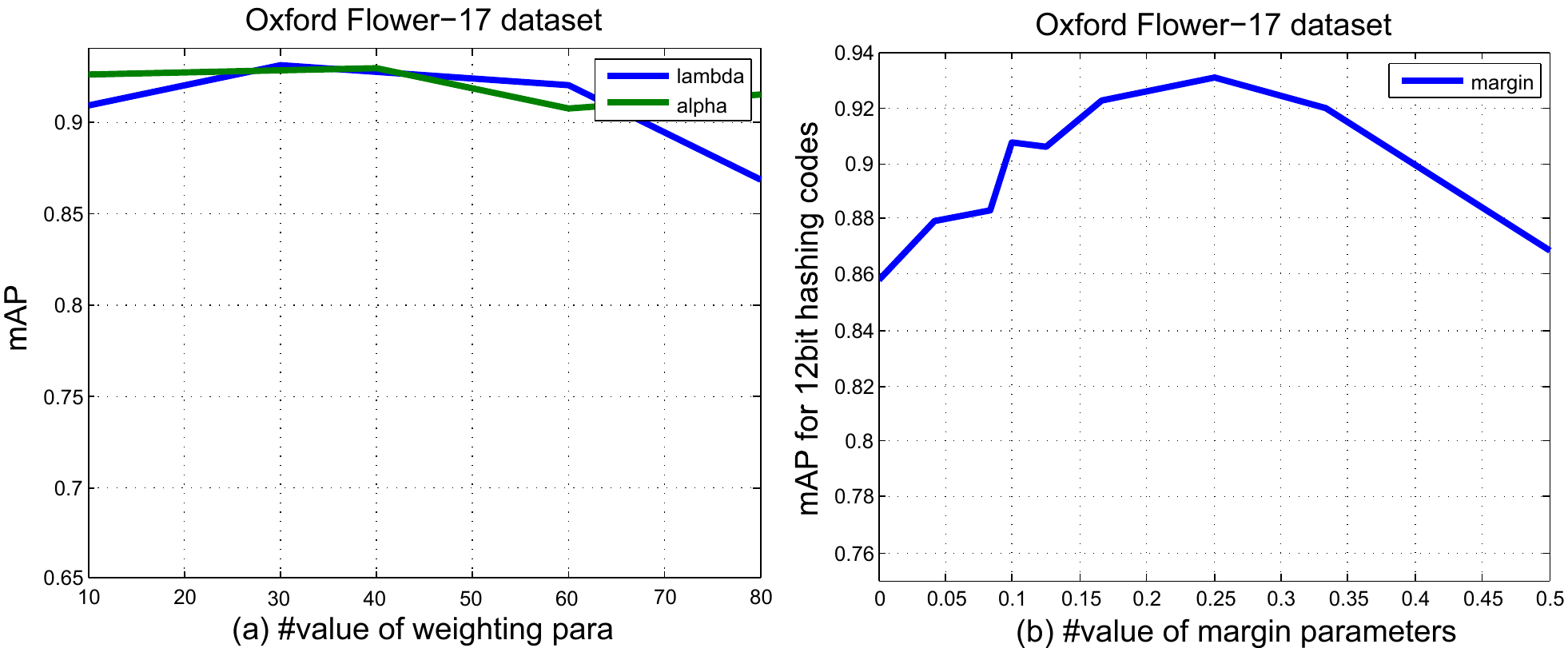}   
    	\end{minipage}%
    	\begin{minipage}[t]{0.33\linewidth}   
    		\centering   
    		\includegraphics[width=1in]{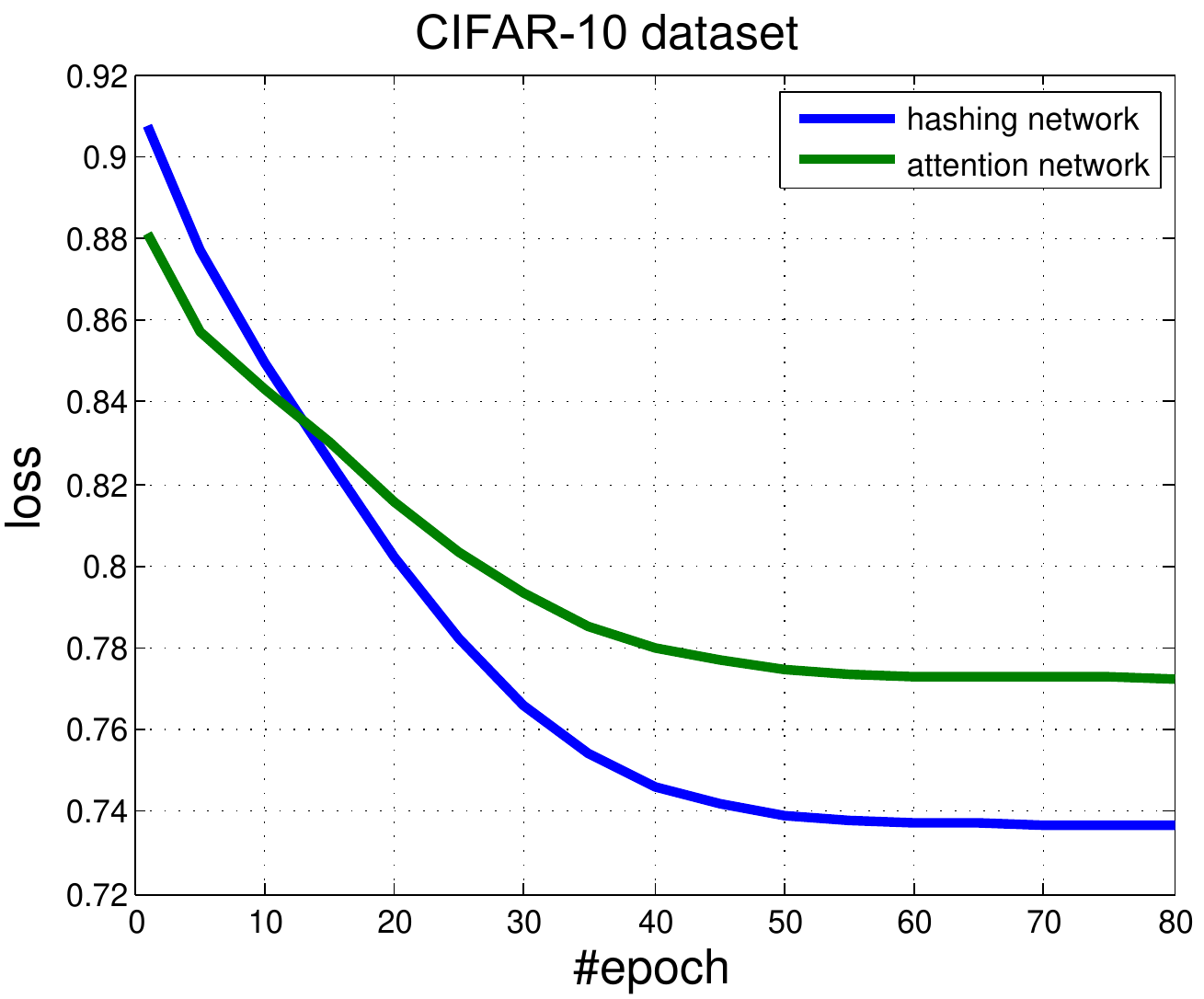}   
    	\end{minipage}   
    	\caption{Sesitiveity to hyper-parameters (a, b) and the convergence of the attention and hashing networks.} 
   	    \label{fig:ana} 	  
    \end{figure}

{\bfseries{Hyper-Parameters Analysis}} In this subsections, we study the effect of the hyper-parameters. The experiments are conducted on Oxford Flower-17. The quantization penalty parameter $\lambda$ and saliency penalty parameter $\alpha$ is selected by cross-validation from 1 to 100 with an additive step-size 10. Fig.~\ref{fig:ana}(a) shows that DSaH is not sensitive to the hyper-parameters $\lambda$ and $\alpha$ in a large range. For example, DSaH can achieve good performance on Oxford-17 with $10 \leqq \lambda \leqslant 80$. As shown in Fig.~\ref{fig:ana}(b), the value of margin parameters $m$ should not be too large or too small. This is because that according to Eq.~\ref{eq:saliencyloss}, if the value of margin is too large, the saliency loss is equal to the semantic loss. If the value of margin is too small, the saliency loss punishes the saliency image to be similar to the original images.


{\bfseries Convergence of Networks} Since our method trains the attention network and the hashing network iteratively, we study the convergence of the proposed networks in CIFAR-10 dataset. As shown in Fig.~\ref{fig:ana}, it can be seen that both the attention network and the hashing network converges after a few epochs, which shows the efficiency of our solution.

{\bfseries Component Analysis of the Loss Function} Our loss function consists of two major components: semantic loss $J_{sem}$ and saliency loss $J_{sal}$. To evaluate the contribution of each loss, we study the effect of different loss combinations on the retrieval performance. The experimental results are shown in TABLE~\ref{lossanalysis}. An interesting observation is that for the attention networks, $J_{sal}$ achieves better performance than $J_{sem}$. The result is understandable because we use the attention networks to highlight the most discriminative regions. Yet the semantic loss only punishes the network so that it can locate the semantic-preserving regions. Another interesting finding is that for the hashing networks, using the combination of $J_{sal}$ and $J_{sem}$ can obtain even worse performance than using $J_{sem}$ only. A possible reason is that when we use the saliency loss for the hashing networks, the binary codes $b'_{i}$ learned from saliency image is required to be more discriminative than $b_{i}$ from the original images. This might force the hashing codes $b_i$ of the original image to become worse and make $b_i$ less effective in guiding the attention network to highlight salient regions. As shown in TABLE~\ref{lossanalysis}, the best performance is achieved when we use the combination of the two components, $J_{sal}$ and $J_{sem}$, for the attention networks and only use the semantic loss $J_{sem}$ for the hashing networks.

%
\begin{table}[htb]
	\small
	\vspace{-0.3cm}
	\centering
	\caption{The \textbf{MAP} of DSaH on Stanford Dog-120 using different combinations of components.}
	\label{lossanalysis}
	\begin{tabular}{r|r|ccc}
		\hline
		\multirow{2}{*}{Hashing-Net}&\multirow{2}{*}{Attention-Net}&
		\multicolumn{3}{c}{Stanford Dog-120}\cr\cline{3-5}
		&&12 bits&24 bits&48 bits\cr
		\hline
		\multirow{3}{*}{$J_{sem}+J_{sal}$}
		&$J_{sem}$&-&-&-\cr\cline{2-5}
		&$J_{sal}$&-&-&-\cr\cline{2-5}
		&$J_{sem}+J_{sal}$&0.3864&0.5032&0.6355\cr\cline{2-5}
		\hline
		\multirow{3}{*}{$J_{sem}$}
		&$J_{sem}$&0.3374&0.4738&0.5931\cr\cline{2-5}
		&$J_{sal}$&0.3756&0.5051&0.6275\cr\cline{2-5}
		&$J_{sem}+J_{sal}$&{\bf 0.3976}&{\bf 0.5283}&{\bf 0.6452}\cr\cline{2-5}
		\hline
	\end{tabular}
\end{table}

\begin{figure}
	\centering
	\includegraphics[width=0.9\linewidth]{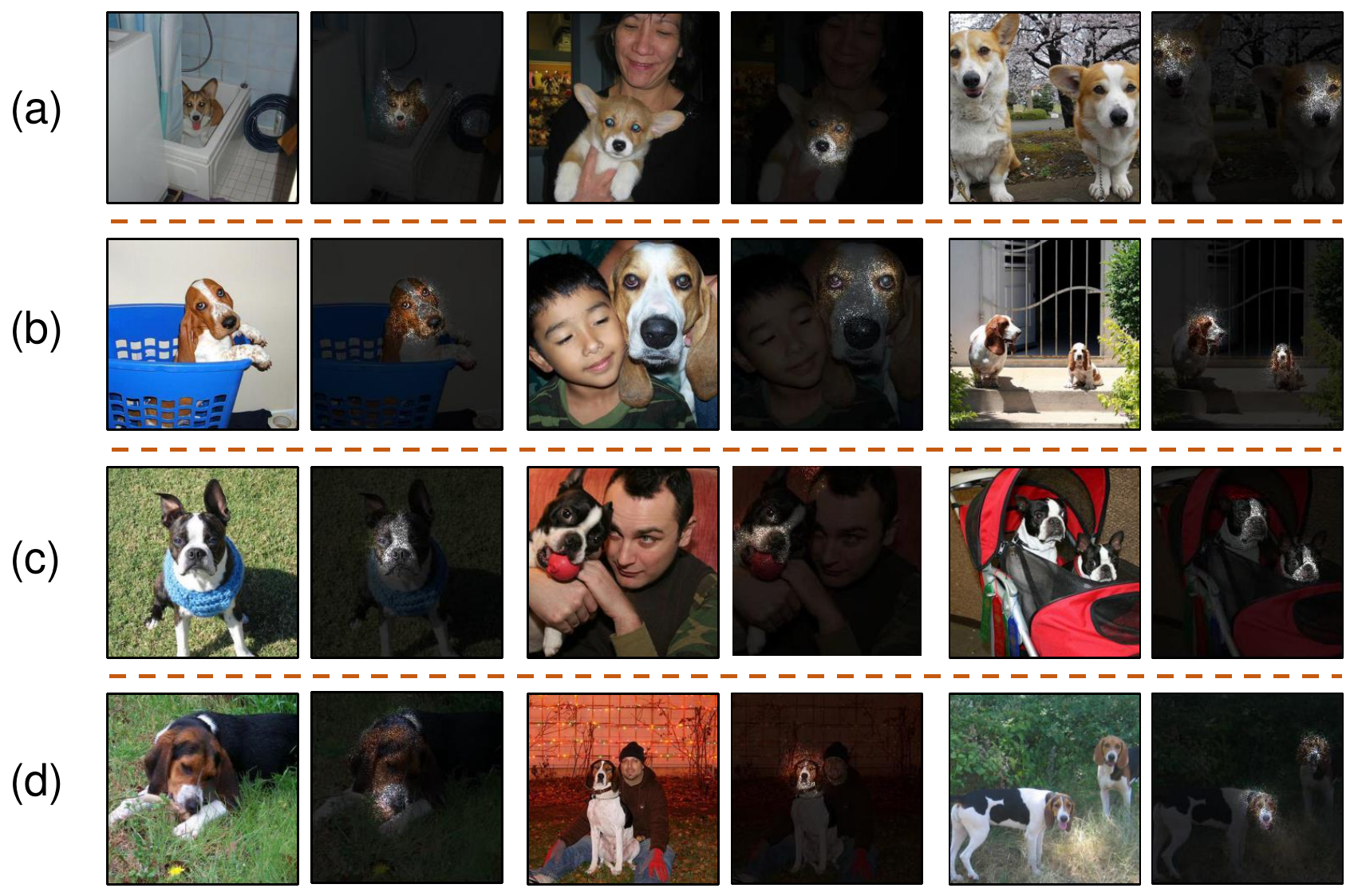}
	\caption{Examples of the salient region learned by the attention network for Stanford Dogs-120 dataset. As the most import part, the heads of dogs are correctly highlighted in the saliency images under various conditions. }
	\label{Fig:Saliency}
	\vspace{-0.55cm}
\end{figure}

{\bfseries Learned Salient Region} Fig.~\ref{Fig:Saliency} shows some typical samples, including multi-objects, occlusion and so on. Each row of Fig.~\ref{Fig:Saliency} is corresponding to a single category. We have several observations about the learned saliency regions. Most importantly, these learned saliency regions always cover the heads of dogs. This is because the head region is important for distinguishing the breed of dog. The typical samples are detailed as: 

(1) The first image in Fig.~\ref{Fig:Saliency}(a) has a complex background. (2) The first image in Fig.~\ref{Fig:Saliency}(b) shows that the body of a dog is overshadowed (3) Compared the first image in Fig.~\ref{Fig:Saliency}(c) with Fig.~\ref{Fig:Saliency}(d), the dogs are in different positions (sitting on or lying on grassland). The head region is accurately mined no matter how the face is oriented (frontal or not). (4) For the second image of each line in Fig.~\ref{Fig:Saliency}, human body was regarded as background. (5) The third image of each line in Fig.~\ref{Fig:Saliency} exists more than one dog. The discriminative region of both dogs could be detected. Compared with the third image Fig.~\ref{Fig:Saliency}(c) and Fig.~\ref{Fig:Saliency}(d), the distance between two dogs does not affect the saliency results. (6) The scales of the objects shown in the first image of Fig.~\ref{Fig:Saliency}(a) and the second image of Fig.~\ref{Fig:Saliency}(b) has a significant difference.  The result shows that the heads of dogs are also correctly highlighted in the saliency images.

\section{Conclusion}
In this paper, we propose a novel supervised deep hashing method for fine-grained retrieval, named deep saliency hashing (\textbf{DSaH}). To distinguish fine-grained objects, our method consists of an attention network to automatically mine discriminative region and a parallel hashing network to learn semantic-preserving hashing codes. We train the attention model and the hashing model alternatively. The attention model is trained based on the semantic loss, quantization loss, and saliency loss. Based on semantic loss and quantization loss, we obtain semantic-preserving hashing codes from the hashing model. Extensive experiments on CIFAR-10 and NUSWIDE dataset demonstrate that our proposed method is comparable to the state-of-art methods for traditional hashing retrieval task. And the experiments on Oxford Flower-17, Stanford Dogs-120 and CUB Bird datasets show that our method achieves a significant improvement for fine-grained retrieval.
	
{\small
\bibliographystyle{ieee}
\bibliography{dsahref}
}

\end{document}